\documentclass[conference]{IEEEtran}
\IEEEoverridecommandlockouts
\usepackage{cite}
\usepackage{amsmath,amssymb,amsfonts}
\usepackage{algorithmic}
\usepackage{graphicx}
\usepackage{textcomp}
\usepackage{xcolor}
\usepackage{hyperref}
\usepackage{nameref}
\usepackage{array}
\usepackage{booktabs}
\usepackage{float}
\usepackage{makecell}
\usepackage{multirow}  
\usepackage[caption=false,font=footnotesize]{subfig}
\def\BibTeX{{\rm B\kern-.05em{\sc i\kern-.025em b}\kern-.08em
    T\kern-.1667em\lower.7ex\hbox{E}\kern-.125emX}}
\begin{document}

\newcolumntype{P}[1]{>{\centering\arraybackslash}p{#1}}

\newcommand{\dataset}{\textsc{UWGB-StreetCrack}}
\newcommand{\maskrcnn}{Mask R-CNN}
\makeatletter
\newcommand{\setlabel}[1]{\edef\@currentlabel{#1}\phantomsection\label}
\makeatother

\newcommand{\hlabel}{\phantomsection\label}
\newcommand{\nazim}[1]{\textcolor{red}{#1}}
\title{Pixel-Level Pavement Distress Assessment Using Instance Segmentation}

\author{
  \IEEEauthorblockN{
    Logan Dewick, Bibesh Pyakurel, Kong Pheng Yang, Sumit Karki,
    Nazim Choudhury, and M. G. Sarwar Murshed
  }
  \IEEEauthorblockA{
    \textit{Computer Science Department} \\
    \textit{University of Wisconsin--Green Bay} \\
    Green Bay, WI, USA \\
    \{dewil01, pyakb27, yangk41, karks03, choudhun, murshedm\}@uwgb.edu
  }
}

\maketitle

\begin{abstract}
Automated pavement distress assessment requires more than image-level classification or coarse bounding box detection, demanding precise localization of thin, branching, and irregular cracks to achieve the geometric precision necessary for maintenance-relevant quantification. This paper presents a vision-based pavement distress analysis system based on \maskrcnn{} instance segmentation and evaluates it on \dataset{}, a custom field-collected roadway image dataset acquired with a vehicle-mounted iPhone 15 Pro Max and manually annotated with polygon labels for longitudinal cracks, transverse cracks, alligator cracks, and potholes. Five Detectron2-based \maskrcnn{} backbone variants were considered under a consistent fine-tuning protocol. The best-performing test result was obtained by \maskrcnn{} with a ResNet-101 FPN backbone, which achieved 84.23\% precision, 90.04\% recall, and an F1 score of 87.04\%. The same model produced an aggregate predicted distress-area fraction of 2.164\%, closely matching the 2.170\% ground-truth distress-area fraction. To contextualize the segmentation system against a detector-oriented alternative, a CSPDarknet53-based YOLO detector was also adapted and retrained on the UWGB data, reaching 27.5\% precision and 20.7\% recall under the object-level evaluation protocol. The results show that instance segmentation is a practical direction for field pavement imagery and aggregate distress-area estimation, while also exposing open challenges in annotation consistency, class imbalance, confounder rejection, and mask-level benchmarking.

\end{abstract}


\section{Introduction}
\label{sec:introduction}
Different types of pavement cracks (e.g., longitudinal, transverse, alligator, block, thermal) signal structural deterioration from traffic loading, thermal cycling, moisture, and material fatigue~\cite{van2002guidelines}.
If left untreated, they escalate maintenance costs and compromise safety.
Deferred roadway maintenance also creates a
substantial economic burden, motivating automated and quantitative pavement-assessment systems that can support maintenance prioritization and infrastructure planning~\cite{pewresearch25}.

Manual pavement inspection remains the dominant practice and is labor-intensive, time-consuming, potentially hazardous, and subject to variability among inspectors and operating conditions~\cite{huang2024application}. 
Deep learning and computer vision can automate roadway inspection and analyze large volumes of field imagery~\cite{mandal2020,du2021pavement,wang2025gsbyolo}. However, the usefulness of these methods depends on the spatial detail provided by their outputs.
Image- and patch-level classification methods determine whether distress is present but do not localize it~\cite{tabernik2023automated,nie2018pavement}.
Object detectors localize defects using bounding boxes~\cite{wang2018road}, but rectangular regions provide limited geometric precision for thin, branching, and irregular cracks. Semantic-segmentation methods provide pixel-level delineation~\cite{lau2020automated,bang2019encoder,feng2020fusion}, but are unable to preserve the identities of individual distress instances in scenes containing multiple or interconnected cracks~\cite{ye2024yolov7wmf}.

Instance segmentation addresses these limitations by jointly predicting the class, location, and pixel-level mask of each detected object.
Mask R-CNN~\cite{He2017MaskRCNN}, for example, combines object detection with a parallel mask-prediction branch, enabling both object-level localization and aggregate area estimation. 
This capability is particularly relevant to full-scene roadway imagery, where multiple distress types may
coexist with visually similar non-distress features such as shadows, painted markings, tire marks, stains, and manhole edges. 
Although pavement-specific instance-segmentation methods have emerged~\cite{ye2024yolov7wmf,wang2023sparseinstcdsm}, evaluation on smartphone-collected, full-scene roadway imagery containing multiple distress categories and realistic field confounders remains limited.

This paper presents an applied empirical evaluation of \maskrcnn{} on \dataset{}, a field-collected roadway dataset with polygon-level annotations for longitudinal cracks, transverse cracks, alligator cracks,
and potholes. 
Rather than introducing a new network architecture, this study
examines the performance and practical limitations of established \maskrcnn{} backbone configurations on challenging roadway images. 
A
CSPDarknet53-based YOLO detector~\cite{mandal2020} and a crop-based segmentation model are also included as contextual reference models.

The main contributions of this study are:
\begin{enumerate}
 \item \dataset{} dataset: smartphone-based roadway imagery with polygon-level annotations for four pavement distress classes. 
 \item \maskrcnn{} pipeline: instance segmentation framework for distress localization, mask prediction, and area estimation.
 \item Detection baseline: CSPDarknet53 YOLO \cite{mandal2020} adapted to \dataset{} for methodological comparison. 
 \item Failure analysis: an analysis of visual confounders, class imbalance, and annotation ambiguity affecting automated pavement
    assessment.
\end{enumerate}
\section{Related Work}
\setlabel{2}{sec:relatedwork}
Most existing pavement instance-segmentation work targets cropped regions, single-crack scenes, or controlled acquisition conditions.
 In contrast, full-scene multi-distress assessment on smartphone-collected roadway imagery remains relatively understudied.
This section reviews four major paradigms in pavement-distress
analysis and identifies the research gap motivating this study.

 Classical and feature-engineered crack detection studies (e.g., CrackTree \cite{zou2012cracktree} and CrackForest  \cite{shi2016crackforest}) relied on thresholding, edge detection, morphology, wavelet analysis, path-based extraction, and other handcrafted image-processing operations. 
Alongside their computational appeal, these methods are sensitive to illumination changes, pavement texture, shadows, stains, and road markings. 
Their dependence on handcrafted assumptions limits transferability across road surfaces and acquisition conditions.

Deep convolutional neural networks (CNN) introduced semantic segmentation and reduced reliance on handcrafted features by learning representations directly from image data. 
Earlier models in this category exploited patch classification \cite{zhang2016cnncrack} or structured crack-probability prediction \cite{fan2018structured}.
In addition, encoder-decoder architectures (e.g., \cite{ronneberger2015unet,bang2019encoder}) and pavement-specific variants (e.g., DeepCrack \cite{zou2018deepcrack}, FPCNet \cite{liu2019fpcnet}) further improved pixel-level delineation via multi-scale fusion.
Despite the pixel-level crack region prediction, semantic segmentation alone generally does not separate adjacent distress instances - a limitation when scenes contain multiple or ambiguously connected cracks.


Object detectors (e.g., Faster R-CNN \cite{ren2015faster, murshed2024ORPN}, YOLO-family models \cite{mandal2020, hu2021deepmodels}, CenterNet, and EfficientDet) are efficient for multi-class localization.
More recent detector-oriented studies \cite{liang2024improvedyolov7,wang2025gsbyolo} have improved speed and robustness through lightweight multi-scale feature fusion and YOLO modifications.
Hybrid pipelines (e.g., \cite{feng2020fusion,liu2020twostep}) in this category have attempted to combine both detection and segmentation. 
These approaches highlight the practical value of segmentation whenever crack geometry or area is needed.

Instance segmentation models \cite{He2017MaskRCNN,lin2017fpn} aim to retain the localization advantages of detectors while producing masks for each detected object. 
Mask R-CNN~\cite{He2017MaskRCNN} extends Faster
R-CNN~\cite{ren2015faster} with a parallel mask-prediction
branch, while Feature Pyramid Networks~\cite{lin2017fpn}
improve multi-scale feature representation.
Pavement-specific instance-segmentation work  \cite{ye2024yolov7wmf, wang2023sparseinstcdsm} has also emerged for real-time crack detection.

Most segmentation studies focus on cropped regions, single-crack scenes, or datasets collected under narrower imaging conditions than those encountered in routine field acquisition. 
The current study is positioned in this gap by evaluating \maskrcnn{} instance segmentation on a custom field-collected roadway dataset with four pavement-distress categories.
\section{Dataset}
\setlabel{3}{sec:dataset}

\subsection{Acquisition Protocol}
The \dataset{} dataset was collected from roadway imagery using an iPhone 15 Pro Max mounted on the front of a vehicle.
Videos were recorded while the vehicle traveled on local roads and were subsequently converted into still frames. 
A Python-based filtering procedure reduced repeated coverage of the same roadway regions. 
Three research assistants then reviewed the extracted frames and removed duplicates, blurred images, and images with insufficient roadway visibility.

\subsection{Annotation Taxonomy and Quality Control}
The cleaned images were annotated in Label Studio (a free and open-source web-based tool for annotating images with machine learning labels) \cite{Label_Studio} using polygon masks.
The four-class taxonomy used in the stored annotation files consists of longitudinal, transverse, alligator cracks, and potholes.
Longitudinal cracks run approximately parallel to the roadway direction, whereas transverse cracks run approximately perpendicular to it. 
Alligator cracks are interconnected crack networks associated with repeated loading or structural failure, and potholes are bowl-shaped depressions often associated with the progression of untreated cracking \cite{miller2003distress}.
Each annotation contained a class label, a bounding box, and one or more polygon segments, which were later converted to COCO-style instance annotations.

Three research assistants performed the annotation after training
with the PASER pavement-rating manual~\cite{Walker}. The manual
provided common visual criteria for distinguishing distress types
and was used to improve annotation consistency. Ambiguous cases
included faint cracks, stains, painted markings, manhole edges,
and interconnected crack networks that could be interpreted as
one alligator-crack instance or several linear cracks. 
\begin{table}[t]
\centering
\caption{Dataset split and class-level instance counts.}
\label{tab:dataset_summary}
\scriptsize
\setlength{\tabcolsep}{2.4pt}
\renewcommand{\arraystretch}{1.15}
\renewcommand\theadfont{\scriptsize\bfseries}
\begin{tabular}{@{}lcc*{4}{c}@{}}
\toprule
& & & \multicolumn{4}{c}{\textbf{Class-level instances}} \\
\cmidrule(lr){4-7}
\thead{Split} &
\thead{Images} &
\thead{Instances} &
\thead{Longi-\\tudinal} &
\thead{Trans-\\verse} &
\thead{Alli-\\gator} &
\thead{Pot-\\hole} \\
\midrule
Train + validation & 1309 & 1484 & 639 & 698 & 82 & 65 \\
Test               & 231  & 261  & 112 & 123 & 14 & 12 \\
\bottomrule
\end{tabular}
\end{table}
Table~\ref{tab:dataset_summary} reports the split and class-level instance
counts. The training and validation partitions contain 1,309
images and 1,484 instances, while the held-out test partition
contains 231 images and 261 instances.
Frames from the same video sequence or roadway segment were
assigned to only one partition to reduce leakage of visually
overlapping roadway contexts.

\section{Methodology}
\begin{figure*}[!t]
    \centering
    \includegraphics[width=0.88\linewidth]{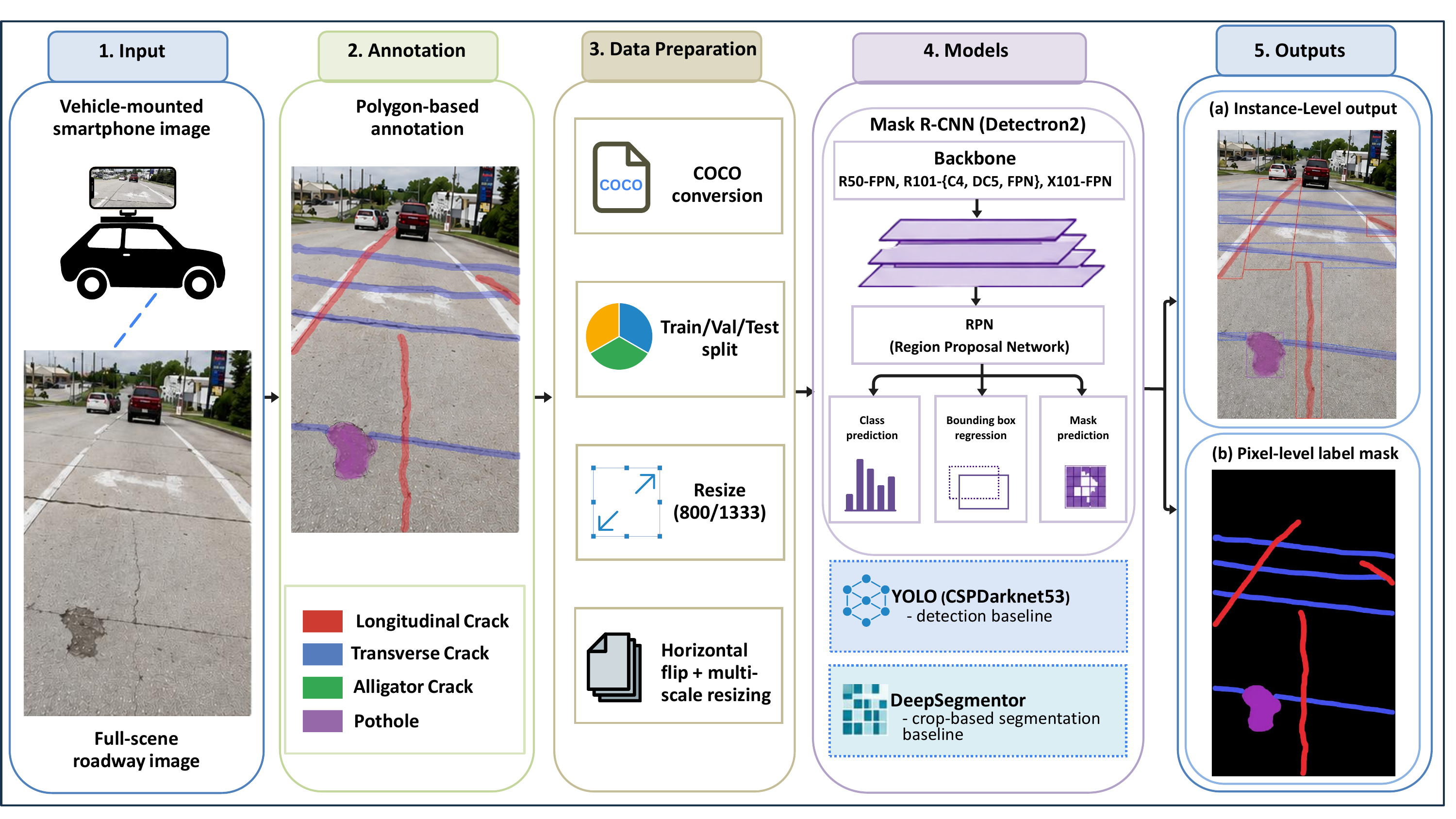}
    \caption{Pavement-distress assessment pipeline from roadway-image
acquisition and polygon annotation to Mask R-CNN-based instance and
pixel-level predictions.}
    \label{fig:overall_pipeline}
\end{figure*}

\setlabel{4}{sec:method}
Our approach leveraged Mask R-CNN implemented in Detectron2, a mature and widely adopted framework for instance segmentation. 
An overview of the complete pipeline is shown in Fig.~\ref{fig:overall_pipeline}. 
We evaluated five backbone variants (ResNet-50, ResNet-101, and ResNeXt-101 with different feature-extraction strategies) using consistent hyperparameters to identify the most effective architecture.
To contextualize our results and demonstrate the advantages of instance segmentation, we established two baseline comparisons: (1) a detection-based baseline using CSPDarknet53-based YOLO \cite{mandal2020}, which localizes objects but cannot produce pixel-level masks, and (2) a segmentation-based baseline using DeepSegmentor \cite{zou2018deepcrack, liu2019deepsegmentor}, which achieves pixel-level precision but only on manually cropped image regions. 
This comparative approach highlights both the strengths of instance segmentation and the specific limitations of prior approaches. 
The following subsections detail the data processing pipeline, architectural choices, training protocols, and inference configuration used throughout.

\subsection{Instance-Segmentation Pipeline}
The current study leveraged the \maskrcnn{} \cite{He2017MaskRCNN} deep learning framework
implemented in Detectron2 \cite{wu2019detectron2}. 
The pipeline consists of four stages: (i) field image curation and annotation, (ii) binary instance masks conversion and data loading, (iii) fine-tuning of \maskrcnn{} variants initialized from COCO (Common Objects in Context)-pretrained checkpoints, and (iv) evaluation.
During training, Label Studio polygon coordinates were interpreted as closed contours and rasterized into binary masks using the standard COCO polygon mechanism available in Detectron2. 
Invalid polygons with fewer than three vertices were excluded.
Each valid polygon generated one instance mask aligned to the image coordinate system.
No offline cropping was used for the full-image \maskrcnn{}.

\subsection{Preprocessing and Augmentation}
All images were resized while preserving the aspect ratio. 
The shorter and the longer sides of the images were resized to 800 and 1333 pixels, respectively, for the selected \maskrcnn{} family.
This process was performed according to the model-zoo protocol of Detectron2.
Zero padding, when needed for batching and stride-compatible tensors, was applied only after resizing.

The conservative augmentation policy was intentional owing to the substantial appearance variation of field images.
Training used random horizontal flipping with probability 0.5 and multi-scale resizing. 
No other augmentation (e.g., color jitter, blur, CutMix, mosaic augmentation, random rotation, or synthetic perturbation) was used in training.

\subsection{Mask R-CNN Variants}
Five \maskrcnn{} backbone variants were considered in the project pipeline. Table~\ref{tab:model_variants} summarizes the model families and their relative backbone complexity. 
Among all the variants (e.g., ResNet-C4, ResNet-DC5, ResNet-FPN) available in Detectron2, R50-FPN and R101-FPN achieved the strongest validation performance and were selected for detailed test-set evaluation.

\begin{table}[!t]
\centering
\caption{Architectural and computational summary of the \maskrcnn{} variants considered in this study. Parameter counts and FLOPs are approximate reference values.}
\label{tab:model_variants}
\renewcommand{\arraystretch}{1.15}
\setlength{\tabcolsep}{4pt}
\small
\begin{tabular}{lccc}
\toprule
\textbf{Variant} & \textbf{Params. (M)} & \textbf{GFLOPs} & \textbf{Detector Cost} \\
\midrule
R50-FPN 3x  & 25.6 & 4.1  & Medium \\
R101-C4     & 44.5 & 7.8  & Med.-high \\
R101-DC5     & 44.5 & $>7.8$ & High \\
R101-FPN 3x & 44.5 & 7.8  & High \\
X101-FPN    & 88.8 & 16.5 & Very high \\
\bottomrule
\end{tabular}
\end{table}

\subsection{Mask R-CNN Training Protocol}
Training and inference were performed on an Ubuntu 24.04 LTS workstation equipped with an NVIDIA GeForce RTX 4080 SUPER GPU, an Intel Core i9 CPU, and 32 GB DDR4 RAM. 
For the reported \maskrcnn{} experiments, the annotated images were split into training, validation, and test sets using a 70/15/15 ratio. The held-out test set consisted of 231 images containing 261 labeled distress instances.
All \maskrcnn{} variants were initialized from COCO-pretrained model-zoo weights and fine-tuned with stochastic gradient descent using momentum 0.9, weight decay 0.0001, initial learning rate 0.001, and global batch size 8. 
Across all evaluated variants, training followed a learning rate decay set to a factor of 0.1 at epoch 24 and epoch 33.
Models were configured to train for up to 100 epochs with early stopping when validation performance no longer improved. Validation performance peaked at epoch 40, and the corresponding checkpoint was used for test-set evaluation.

\subsection{Baseline Detector Evaluation}

A CSPDarknet53-based YOLO detector~\cite{mandal2020} was adapted to the four-class \dataset{} dataset. Unlike Mask R-CNN, this detector predicts bounding boxes but does not generate pixel-level masks. The polygon annotations were therefore converted to normalized YOLO bounding-box format.
The detector was trained from scratch using an input resolution of $640 \times 640$, a batch size of 16, the default optimizer configuration, and 100 epochs. Training generally converged after 70--80 epochs, and extending training to 200 epochs did not yield significant performance improvements.
The result of the best-performing model is reported in Section~\ref{sec:results}.

\subsection{Baseline Segmentation-Based Evaluation}
In addition, we also considered another baseline segmentation-based model named DeepSegmentor - a CNN-based pixel-level distress segmentation approach originally designed to predict distress regions from image patches.
Unlike the CSPDarknet53-based YOLO model and the current study, DeepSegmentor is incapable of processing full roadway images, however, it is capable of crop-level segmentation.
Therefore, distress regions were manually cropped from full roadway images for pixel-level prediction. 
The resulting masks were used to compute the detected distress-area percentage and compare it with the ground-truth annotated distress area. 

It is noteworthy that, unlike the two baseline models, the proposed Mask R-CNN framework is an end-to-end model that performs simultaneous localization, classification, and pixel-level segmentation on full roadway images.

\subsection{Inference Configuration}
At inference time, the same training image properties (i.e., aspect ratio, size) were used without stochastic augmentation. 
The Region Proposal Network (RPN) used anchor sizes of 32, 64, 128, 256, and 512 pixels and aspect ratios of 0.5, 1.0, and 2.0. 
The RPN non-maximum suppression threshold was 0.65, and the final detection-stage non-maximum suppression threshold was 0.5. 
The final confidence threshold was set to 0.75 on the validation partition because it provided the best precision-recall balance; it was then held fixed for test evaluation.

\section{Evaluation Protocol}
For the evaluation purpose, we compared our results in both object-detection (object-level) and crack segmentation (pixel-level).
The \maskrcnn{} precision, recall, and F1 values should therefore be interpreted as object-level performance.
Predicted objects were matched to the ground-truth objects by considering bounding-box Intersection over Union (IoU).
A prediction was counted as a true positive if its class label matched the ground-truth class and its bounding-box IoU was at least 0.1. 
Each predicted instance could be assigned to at most one ground-truth instance. 
Unmatched predictions were counted as false positives, and unmatched ground-truth instances were counted as false negatives.


The detected distress area was computed at the pixel level by summing predicted or annotated mask pixels and normalizing by the total image area across the evaluated set.
\section{Results} \label{sec:results}

Table~\ref{tab:comprehensive_results} summarizes the performance values for the evaluated pavement-crack analysis models on the \dataset{} dataset.
The results are reported using precision, recall, F1 score, and the percentage of detected crack area. 



\begin{table}[t]
\centering
\caption{Comprehensive performance comparison of models on the pavement crack dataset.}
\label{tab:comprehensive_results}

\scriptsize
\setlength{\tabcolsep}{2.5pt}
\renewcommand{\arraystretch}{1.15}
\renewcommand\theadfont{\scriptsize\bfseries}

\begin{tabular}{@{}lcccc@{}}
\toprule
\thead{Model} &
\thead{Precision\\(\%)} &
\thead{Recall\\(\%)} &
\thead{F1\\(\%)} &
\thead{Aggregate \\distress area (\%)} \\
\midrule

Ground truth
& -- & -- & -- & 2.170 \\

\makecell[l]{Mask R-CNN ResNet-50\\FPN $3\times$}
& 75.56 & 89.01 & 81.73 & -- \\

\makecell[l]{Mask R-CNN ResNet-101\\FPN $3\times$}
& \textbf{84.23}
& \textbf{90.04}
& \textbf{87.04}
& \textbf{2.164} \\

\makecell[l]{YOLO (CSPDarknet53)\\\cite{mandal2020}}
& 27.5 & 20.7 & 23.62 & -- \\

\makecell[l]{DeepSegmentor\\
\cite{zou2018deepcrack,liu2019deepsegmentor}}
& -- & -- & -- & 2.130 \\

\bottomrule
\end{tabular}
\end{table}

\subsection{Model Performance}
Among the evaluated Mask R-CNN variants, the ResNet-101 FPN 3x backbone achieved the best overall detection performance. 
It obtained a precision of 84.23\%, a recall of 90.04\%, and an F1 score of 87.04\%. 
In comparison, the ResNet-50 FPN 3x variant achieved 75.56\% precision, 89.01\% recall, and an F1 score of 81.73\%. 
The deeper ResNet-101 backbone therefore improved precision by 8.67 percentage points and F1 score by 5.31 percentage points, while also providing a modest improvement in recall.
This indicates that the higher-capacity ResNet-101 FPN backbone was more effective at suppressing false detections while maintaining strong crack-detection coverage.
\subsection{Model Comparison}
\subsubsection{Object-level detection}
Compared with the YOLO-based CSPDarknet53 detector, the proposed Mask R-CNN ResNet-101 FPN 3x model showed substantially higher performance values across all three evaluation metrics. 
CSPDarknet53 obtained 27.5\% precision, 20.7\% recall, and an F1 score of 23.62\%, whereas the proposed Mask R-CNN ResNet-101 FPN 3x model achieved 84.23\% precision, 90.04\% recall, and 87.04\% F1 score.
It is observable that the current study outperformed the CSPDarknet53 by 56.73 percentage points in precision, 69.34 percentage points in recall, and 63.42 percentage points in F1-score.
\subsubsection{Pixel-level segmentation}
The detected-area results further highlight the practical advantage of the proposed instance-segmentation framework for pavement-distress assessment. 
The ground-truth distress area accounted for 2.170\% of the evaluated image area. 
The proposed Mask R-CNN ResNet-101 FPN 3x model estimated 2.164\% distress area, producing a trivial difference (i.e., 0.006 percentage points) from the ground truth.
In comparison, DeepSegmentor estimated 2.130\% detected-area where the difference (i.e., 0.040 percentage points) is larger than the proposed model in the current study.
Note that DeepSegmentor can only be evaluated as a crop-level segmentation method that requires input images to be manually cropped around crack-containing regions.
When applied directly to full roadway scenes, its performance is limited by real-world visual confounders like shadows, pavement texture, stains, road markings, lane lines, and other non-crack objects. 
In contrast, the proposed Mask R-CNN-based framework operates on full roadway images: it first localizes candidate distress regions and then predicts pixel-level masks for the detected instances. 
Therefore, the proposed model not only produced a closer aggregate crack-area estimate than DeepSegmentor under the reported evaluation, but also provided a more deployable end-to-end solution for realistic pavement distress imagery.


\subsection{Qualitative Error Analysis}
\begin{figure}[!hbpt]
    \centering
    \subfloat[Paint marking false positive.\label{fig:paint_fp}]{%
        \includegraphics[width=0.48\columnwidth]{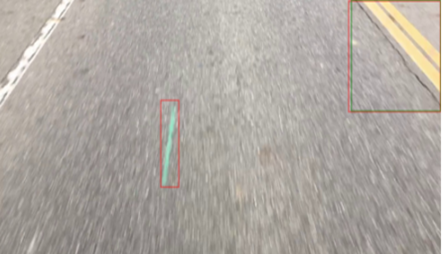}%
    }
    \hfill
    \subfloat[Oil stain false positive.\label{fig:oil_fp}]{%
        \includegraphics[width=0.48\columnwidth]{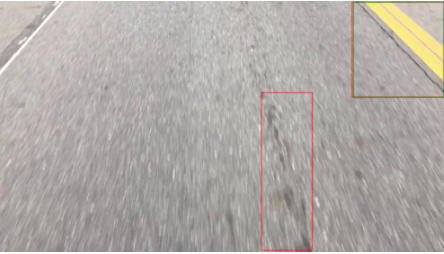}%
    }
    \vspace{4pt}
    \subfloat[Manhole-cover ambiguity.\label{fig:manhole_case}]{%
        \includegraphics[width=0.48\columnwidth]{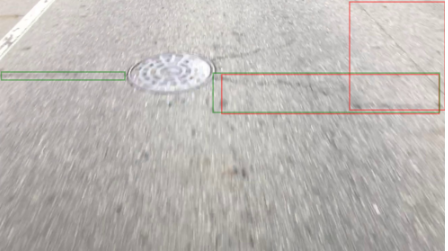}%
    }
    \hfill
    \subfloat[Instance-granularity ambiguity in alligator cracking.\label{fig:alligator_case}]{%
        \includegraphics[width=0.48\columnwidth]{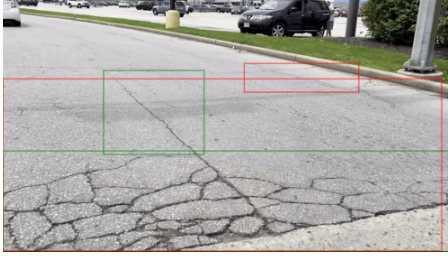}%
    }
    \caption{Representative failure cases used for qualitative analysis. Red boxes denote model predictions and green boxes denote ground-truth annotations.}
    \label{fig:qualitative_cases}
\end{figure}

\autoref{fig:qualitative_cases} shows representative failure cases. 
Red boxes denote model predictions and green boxes denote ground-truth annotations. 
Painted road markings and oil stains produced false positives
because their elongated or high-contrast appearances resembled
cracks.
Manhole-cover scenes were especially ambiguous because faint linear structures adjacent to the cover could be interpreted as either pavement distress or non-distress artifacts. 
Alligator cracks introduced a different difficulty: the model sometimes predicted one connected distress region while the ground truth separated the same region into multiple instances. 
For example, in Figure \ref{fig:alligator_case}, the prediction groups a connected crack network into one instance, whereas the ground-truth annotation separates portions of the same structure into multiple instances. 
This case illustrates ambiguity in defining instance boundaries for
interconnected alligator cracking.
However, the model was successful in correctly predicting the complete distress area (large red bounding box).

These examples highlight why a segmentation benchmark must pair model development with rigorous annotation policy. Without strict guidance, annotators may draw different masks for the same visual structure, especially when a crack network can be interpreted as either one alligator-crack instance or multiple linear cracks. Such ambiguity affects model performance.
\section{Conclusion}
This paper presented a Mask R-CNN framework for automated pavement distress assessment using \dataset{}, a field-collected dataset of 1,540 smartphone roadway images with polygon-level annotations for four distress types. Instance segmentation combines the geometric precision of semantic segmentation with the ability to distinguish multiple distress instances, enabling maintenance-relevant quantification. The best-performing ResNet-101 FPN model achieved 84.23\% precision, 90.04\% recall, and an F1 score of 87.04\%. Its estimated aggregate distress area of 2.164\% closely matched the ground-truth value of 2.170\%, demonstrating its suitability for area-based maintenance decisions. In comparison, CSPDarknet53 YOLO achieved only 27.5\% precision and 20.7\% recall, while the crop-based DeepSegmentor required manual preprocessing and produced a larger area-estimation error of 0.040 percentage points. A key limitation is annotation ambiguity, particularly for alligator cracking, where disagreement over instance boundaries affected model performance. Nevertheless, the results establish instance segmentation as a practical foundation for full-scene pavement distress detection and crack-area estimation, supporting the transition from subjective visual inspection to objective, automated, and evidence-based pavement assessment.

\section*{Acknowledgment}
This work was supported by WiSys through a Spark grant (Award No. 102-24600-4-AAN5138). The authors also thank the University of Wisconsin--Green Bay and the student researchers who contributed to data collection and annotation for \dataset{}.

\section*{Data Governance Note}
The \dataset{} dataset supporting this study is not yet publicly released, as it remains in active use for ongoing research. A public release is planned following further publications and a privacy review of the imagery.

{\footnotesize \bibliographystyle{unsrt}
 \bibliography{references}
 }

\end{document}